\documentclass[final,3p]{elsarticle}
\usepackage{amsmath}
\usepackage[colorlinks=true,breaklinks=true,pdftex]{hyperref}
\usepackage{bbding}
\usepackage{pifont}
\biboptions{authoryear}
\usepackage{xcolor}

\usepackage{amsfonts}
\usepackage[utf8]{inputenc}
\DeclareUnicodeCharacter{202F}{\,}

\begin{document}

\begin{frontmatter}

\title{SSFlowNet: Semi-supervised Scene Flow Estimation On Point Clouds With Pseudo Label}

\author{Jingze Chen$^{1}$, Junfeng Yao$^{*,1,2}$ , Qiqin Lin$^{1}$, Rongzhou Zhou$^{1}$,Lei Li$^{3}$}
\address{$^{1}$Center for Digital Media Computing,School of Film,School of Informatics, Xiamen University.\\
$^{2}$Institute of Artificial Intelligence, Xiamen University.\\
$^{3}$ Department of Computer Science, University of Copenhagen \\
}


\begin{abstract}
In the domain of supervised scene flow estimation, the process of manual labeling is both time-intensive and financially demanding. This paper introduces SSFlowNet, a semi-supervised approach for scene flow estimation, that utilizes a blend of labeled and unlabeled data, optimizing the balance between the cost of labeling and the precision of model training. SSFlowNet stands out through its innovative use of pseudo-labels, mainly reducing the dependency on extensively labeled datasets while maintaining high model accuracy. The core of our model is its emphasis on the intricate geometric structures of point clouds, both locally and globally, coupled with a novel spatial memory feature. This feature is adept at learning the geometric relationships between points over sequential time frames. By identifying similarities between labeled and unlabeled points, SSFlowNet dynamically constructs a correlation matrix to evaluate scene flow dependencies at individual points level. Furthermore, the integration of a flow consistency module within SSFlowNet enhances its capability to consistently estimate flow, an essential aspect for analyzing dynamic scenes. Empirical results demonstrate that SSFlowNet surpasses existing methods in pseudo-label generation and shows adaptability across varying data volumes. Moreover, our semi-supervised training technique yields promising outcomes even with different smaller ratio labeled data, marking a substantial advancement in the field of scene flow estimation.
\end{abstract}


\end{frontmatter}


\section{Introduction}

 \begin{figure}[htb]

  \centering
  \centerline{\includegraphics[width=0.8\textwidth]{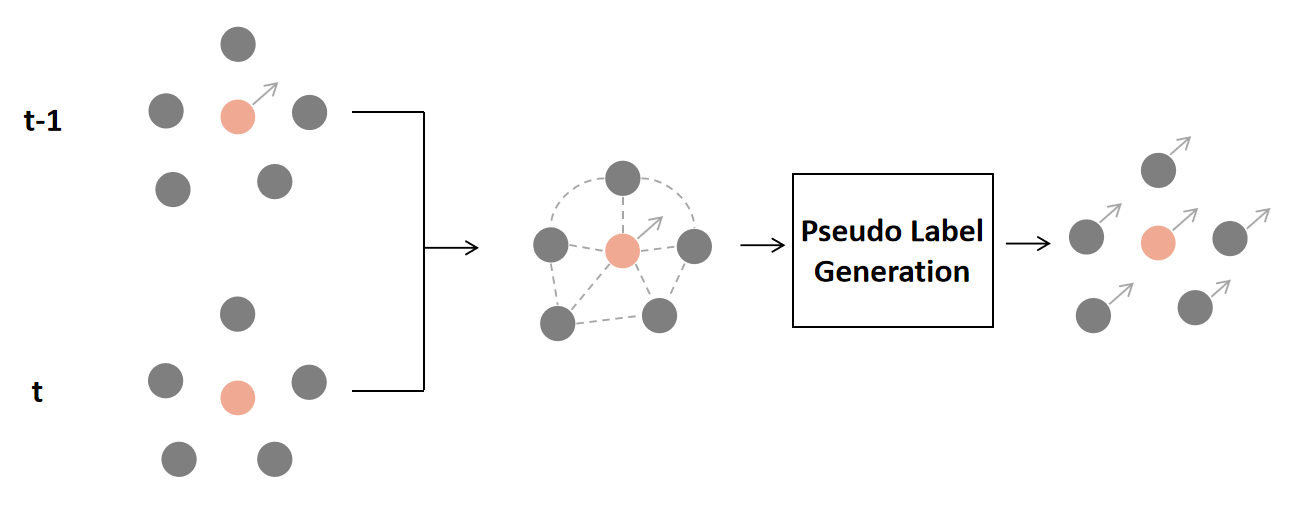}}
%

\caption{The SSFlowNet model dynamically learns point-level similarities by aggregating geometric features across two frames, utilizing global pseudo-labels derived from known true labels to enhance training efficiency.}
\label{fig:res}
\end{figure}



Scene flow, the endeavor to estimate a dense three-dimensional motion field in dynamic environments, exceeds the concept of optical flow by venturing beyond mere two-dimensional motion vector analysis in consecutive image frames. Unlike optical flow, scene flow encompasses a comprehensive capture of the three-dimensional motions of objects and their surrounding environment. This progression from two to three dimensions significantly augments our perception and understanding of dynamic 3D worlds, thereby facilitating a more nuanced acquisition of spatial and temporal information. Consequently, scene flow estimation has become a fundamental tool in many 3D applications, most notably in autonomous driving (AD) and Simultaneous Localization and Mapping (SLAM).

The emergence of deep neural networks has catalyzed advancements in the field of scene flow estimation. A notable milestone was achieved with the introduction of FlowNet3D \cite{liu2019flownet3d}, a model that leveraged the principles established by PointNet++ \cite{qi2017pointnet++} to integrate local features into scene flow estimation, thereby pioneering the application of neural networks in this domain. Following this, HPLFlowNet \cite{Gu_Wang_Wu_Lee_Wang_2019} was developed, featuring an innovative approach that computes multi-scale correlations utilizing upsampling operations within bilateral convolutional layers, thus enhancing the model's depth perception and contextual understanding. Further advancements were made by \cite{Kittenplon_Eldar_Raviv_2020}, which introduced a novel technique for learning a single step of an unrolled iterative alignededment procedure, thereby refining the accuracy of scene flow predictions. Additionally, the advent of 3DFlow \cite{wang2022matters} marked a new era in the field, setting unprecedented benchmarks in both 3D End Point Error (EPE3D) and overall accuracy, thereby exemplifying the dynamic and continuous progression in scene flow estimation methodologies.


Although some work has improved metrics for scene flow estimation, obtaining scene flow labels from the real world can be challenging. \cite{Mittal_Okorn_Held_2019} firstly utilizes an unsupervised method through nearest neighbor loss and cycle consistency loss. However, it ignores the local geometrical properties of point clouds. Inspired by unsupervised methods with point cloud, \cite{Wu_Wang_Li_Liu_Fuxin_2019} further improved the network performance in a coarse-to-fine fashion. However, the errors may accumulate in the early step. \cite{li2021self} proposes a new training mode by pseudo labels, initializing the scene flow through optimal transmission and taking the random walk approach to further improve the label quality. In addition, the pseudo-label-based training approach is also used to address the problem of insufficient label volume. \cite{li2021self} utilized an assignment matrix to guide the generation of pseudo ground truth. Meanwhile, a random walk module was introduced to encourage the local consistency of the pseudo labels. \cite{li2022learning} leveraged monocular images to generate pseudo labels for point clouds, which facilitates the training of scene flow networks. Pseudo-label based training approaches rely on the quality of generated pseudo-labels. Although the pseudo-label based approach has made some progress, the accuracy of pseudo-labels still needs to be improved.

Occupying a unique position in the learning paradigm spectrum, semi-supervised learning (SSL) adeptly mediates the interplay between model accuracy and the magnitude of data. This paper introduces an innovative pseudo-label generation model, a noteworthy advancement in the SSL domain. Grounded in the principles delineated by \cite{Puy_Boulch_Marlet_2020}, our methodology harnesses the capabilities of Graph Neural Networks (GNN) to construct a geometric features graph. This graph is pivotal in developing our correlation matrix, which is meticulously designed to assess the similarities between labeled and unlabeled data points with enhanced precision.

Instead, we integrate a spatial memory module, an innovative addition aimed at fully exploiting the neighboring information. This module enables the construction of a comprehensive graph that interlinks two sequential frames, thereby enriching the model’s understanding of spatial dynamics. To ensure the integrity of flow consistency and the accuracy of the generated pseudo-labels, we have undertaken a significant modification of the existing loss function. This tailored refinement is specifically engineered to better aligneded with the nuanced requirements of our task, thereby bolstering the model’s effectiveness in the complex landscape of semi-supervised learning.

In our empirical study, the model was subjected to extensive evaluation across varied data volumes, precisely at fractions of 1/8, 1/16, 1/32, and 1/64 of the original dataset sizes. We meticulously selected two prominent datasets for this assessment: FlyingThings3D and KITTI. The empirical outcomes reveal that our model exhibits remarkable adaptability and efficacy in conjunction with a spectrum of leading prediction models. This is particularly pronounced in contexts with sparse labeling, where our model consistently demonstrates an exceptional prowess in accurately predicting movement vectors for unlabeled data. Comparatively, when juxtaposed with alternative pseudo-label generation algorithms, the labels produced by our model are discernibly of superior quality. Additionally, the incorporation of our innovative spatial memory module and the association matrix framework has yielded a demonstrable positive impact on the process of label generation, thus significantly enhancing the overall model performance.




The key contributions of our study are outlined as follows:

\begin{itemize}
  \item We have introduced a new semi-supervised learning (SSL) methodology to the field of scene flow estimation. This method effectively reduces the data requirements with only minor compromises in precision.
  \item Our research proposes a correlation matrix based on geometric structures. This matrix is adept at accurately measuring the similarity between distinct points. 
  \item A new flow consistency loss function has been integrated into our model. This enhancement is designed to augment the model's capability for maintaining dynamic flow estimation consistency, which in turn contributes to the improvement of label quality.
\end{itemize}

\section{Related Work}
\textbf{Scene Flow Estimation:} The concept of scene flow was first articulated by Vedula et al. in \cite{vedula1999three}. Scene flow, distinct from the 2D optical flow that delineates the movement trajectories of image pixels, is conceptualized as a vector characterizing the motion of three-dimensional objects. Early research in this field \cite{huguet2007variational, menze2015object, vogel20113d} predominantly utilized RGB data. Notably, Huguet and Devernay \cite{huguet2007variational} introduced a variational approach to estimate scene flow from stereo sequences, while Vogel et al. \cite{vogel20113d} presented a piece-wise rigid scene model for 3D flow estimation. Menze and Geiger \cite{menze2015object} advanced the field by proposing an object-level scene flow estimation method, alongside introducing a dataset specifically for 3D scene flow.

The advent of deep learning heralded transformative approaches in scene flow estimation. PointNet \cite{qi2017pointnet}, as a pioneering work, utilized convolutional operations for point cloud feature learning, which was further refined by PointNet++ \cite{qi2017pointnet++} through feature extraction from local domains. Shi and Rajkumar's Point-GNN \cite{shi2020point} introduced a graph structure to mitigate translation variance. Building on these foundations, subsequent studies \cite{liu2019flownet3d, Puy_Boulch_Marlet_2020, Wu_Wang_Li_Liu_Fuxin_2019,wang2022matters,li2023edge} have achieved impressive results in scene flow estimation. FlowNet3D \cite{liu2019flownet3d}, for instance, leverages PointNet++ \cite{qi2017pointnet++} for feature extraction and introduces a flow embedding layer to capture and propagate correlations between point clouds for flow estimation. Puy et al. \cite{Puy_Boulch_Marlet_2020} employed optimal transport for constructing point matches between sequences. Wu et al. \cite{Wu_Wang_Li_Liu_Fuxin_2019} proposed a cost volume module for processing large motions in 3D point clouds, while Wang et al. \cite{wang2022matters} innovated an all-to-all flow embedding layer with backward reliability validation to address consistency issues in initial scene flow estimation.

The challenge of acquiring accurate ground truth labels for model training has recently led to the exploration of unsupervised strategies. Mittal et al. \cite{Mittal_Okorn_Held_2019} pioneered this approach in scene flow estimation, employing nearest neighbor loss and cycle consistency loss. Li et al. \cite{li2021self} introduced a pseudo-label generation module that synergizes with the methodology of Mittal et al. \cite{Mittal_Okorn_Held_2019}. Furthermore, Li et al. \cite{li2022rigidflow} integrated a local rigidity prior into self-supervised scene flow learning, based on the premise of scenes comprising multiple rigidly moving parts. Ouyang and Raviv \cite{ouyang2021occlusion} incorporated an occlusion mask into their data, training the network to detect occlusions more effectively.


\begin{figure*}
    \centering
  \centerline{\includegraphics[width=16cm]{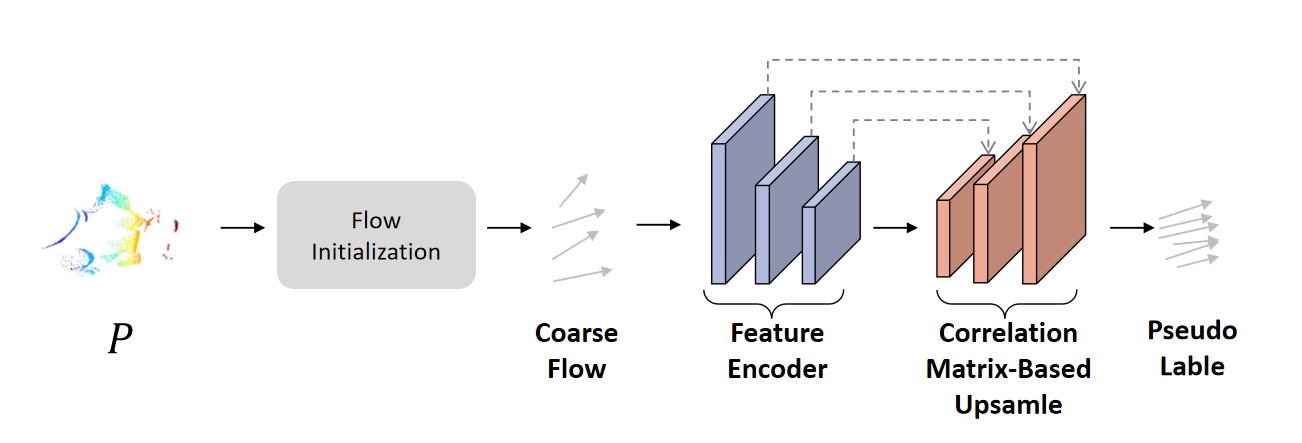}}
 \caption{The overview of proposed SSLFlownet. (i) We start by simply up-sampling the ground truth labels to obtain the coarse flow. (ii) Subsequently, we extract neighboring features at time $t-1$ and assimilate the features at time $t$ through the application of the spatial memory module. (iii) Finally fuse and sum the points based on their features and input to MLP to compute the correlation matrix to generate the pseudo labels.}
 \label{fig1}
\end{figure*}

\textbf{Pseudo-Label Training:} Contemporary advancements in computer vision heavily rely on deep neural networks. A significant limitation of these networks, however, is their reliance on extensive datasets, typically labeled by human annotators or generated automatically through proxy tasks. Lee et al. \cite{lee2013pseudo} introduced a method for predicting pseudo-labels of unlabeled data by analyzing relationships within dataset samples. In this method, for data without labels, the network selects the class with the maximum predicted probability, a technique analogous to Entropy Regularization. Iscen et al. \cite{iscen2019label} furthered this approach by introducing label propagation, utilizing existing ground truth labels to generate new labels.

In the realm of scene flow estimation, several methods based on pseudo-labels have emerged \cite{shen2023self, dong2022exploiting, li2022learning}. Shen et al. \cite{shen2023self} employed a self-supervised approach using pseudo-labels, drawing inspiration from Puy et al. \cite{Puy_Boulch_Marlet_2020} to utilize optimal transport for initializing flow and a random walk technique for label completion. Dong et al. \cite{dong2022exploiting} proposed a method integrating flow into multi rigid-body motion for higher-level scene abstraction. In contrast, Li et al. \cite{li2022learning} developed a multi-modality framework combining RGB images and point clouds to generate pseudo labels for training networks.

Distinct from these methodologies, our model employs label propagation by up-sampling ground truth labels through the learning of geometric features from both labeled and unlabeled data. This approach proves more effective in generating valid labels, thereby facilitating learning in scenarios where labels are sparse or nonexistent.

\textbf{Memory in Scene Flow Estimation:} Memory modules have become increasingly prevalent in 2D video semantic segmentation frameworks. Traditional semantic segmentation on single frames often overlooks the temporal dimension, leading to inaccuracies, particularly in the context of occlusion. Paul et al. \cite{paul2021local} innovatively proposed the use of local Memory Attention Networks, which effectively transform existing single-frame semantic segmentation models into robust video semantic segmentation frameworks. Similarly, Li et al. \cite{li2023memoryseg} introduced a framework for semantic segmentation of temporal sequences in point clouds, utilizing a memory network for the storage and retrieval of past information.

In scene flow estimation, the primary objective is to ascertain motion vectors between two frames, necessitating a deep understanding of the contextual scene information. While there are existing methodologies focused on matching points by learning features between two frames, these often lack in extracting geometric features of points at distinct moments. Our work addresses this gap by introducing a spatial memory module that integrates neighborhood information from different temporal points into a feature encoder. This approach, by incorporating the memory module, adds a temporal dimension to the single-frame feature encoding process. Consequently, this enhancement significantly aids in comprehending contextual features and in constructing a more stable relationship matrix.

\section{Method}

In this section, we formulate the semi-supervised learning problem and then we illustrate our label generation structure as shown in Figure. \ref{fig1}. We first use simple flow up-sampling to get original flow and capture current and future geometric features by spatial memory module. Then we use correlation matrix framework to further refine original flow, finally we illustrate the way in which we maintain flow consistency by imposing penalties on biased points. 
\label{sec:format}

\subsection{Problem Definition}

Consider two point clouds of the same scene captured at two consecutive time instances, denoted as \( P\in \{ p_i \in \mathbb{R}^{3} | i = 1,2, \ldots, n \} \) and \( Q\in \{ q_j \in \mathbb{R}^{3} | j = 1,2, \ldots, n \} \), where \( p_i \) and \( q_j \) represent the \( xyz \) coordinates of the \( i^{th} \) and \( j^{th} \) points in \( P \) and \( Q \), respectively. Let \( F\in \{ f_i \in \mathbb{R}^{3} | i = 1,2, \ldots, n \} \) signify the motion vector corresponding to these two point clouds.

The point set \( P \) is partitioned into two subsets: \( P^l \) and \( P^u \), representing the labeled and unlabeled point sets, respectively. The labeled flow within \( F \) is denoted by \( \hat{F} \), with \( \hat{N} \) labels (\( \hat{N}<N \)). Our pseudo-label generation module aims to estimate the ground truth labels for training purposes. This is articulated in Equation \ref{eq1}, where \( g \) symbolizes the label generation function:
\begin{equation}
    F = g(P, Q, \hat{F})
    \label{eq1}
\end{equation}

\begin{figure*}
    \centering
    \includegraphics[width=16cm]{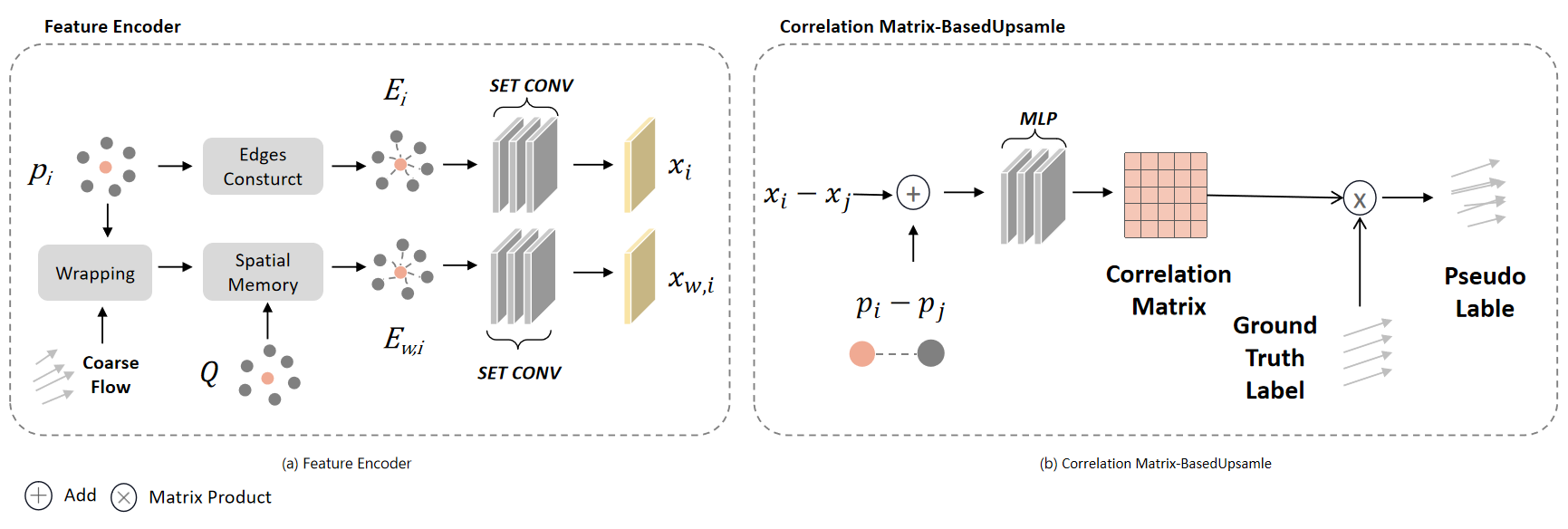}
    \caption{Details about the feature encoder and on-stream sampling. (a) Our feature encoder is divided into two parts: the first extracts the current features, while the second part utilizes spatial memory to save the point features for the subsequent moment and then re-extracts these features. Subsequently, these edge features are processed through the set convolutional layers. (b) The input to the MLP includes the difference in point coordinates as well as their features, with the output being the correlation matrix. This matrix represents the influence factor of each labeled point on the \( p_i \) points.}
    \label{fig2}
\end{figure*}

\subsection{Flow Initialization}
We start by simply up-sampling the flow, the purpose is to obtain original flow and input wrapped points to spatial memory module. To make a first approximation of the global flow according to the known labels $\hat{F}$, we use Euclidean distance to measure the relationship between different points which is used in flow up-sampling. For each unlabelled point $p^l_i$, we select the K nearest labelled points using K-NearestNeighbor(KNN) and perform a weighted average of the flow vectors of these points. This step serves to apply the wrapped points to the spatial memory module.
\begin{equation}
\begin{aligned}
     &D(p,p^l_i) = \|p - p^l_i\|_2 \\
     f_{c,i} &= \sum_{p \in N_i} \frac{\frac{1}{D(p,p^l_i)}}{\sum_{p \in N_i}\frac{1}{D(p,p^l_i)}}
\end{aligned}
\label{eq2}
\end{equation}

$D$ denotes the Euclidean distance function to measure of the straight-line distance between two points in Euclidean space as shown in Figure. \ref{fig2}. $f_{c,i}$ represents the coarse flow after simple up-sampling.
\subsection{Correlation Matrix Guided Label Generation}
Our method attempts to generate pseudo-labels that satisfy the following requirements: (1) Neighbouring points are with similar flow patterns; (2) Points with similar characteristics share similar flow. Thus, we introduce graph-structure based coding on the point cloud to extract point cloud features. First, we construct the graph structure at the point level, where the edges of the graph aggregate information such as the position of the points, colour and normal vector. Then we use setconv layer as encoder to learn neighbourhood information, and subsequently we use MLP to learn the similarity of different point features. 
\subsubsection{Flow-Graph Encoder}
Most current point cloud feature encoder follow the set abstraction layers of PointNet++ \cite{qi2017pointnet++} called Flow-Graph encoder, which obtains point cloud context by local grouping and max pooling.But it fails to utilize the spatial distribution of the input point cloud. This is because hierarchical feature learning fails to encode the spatial distribution in the division of the point clouds.


Thus, we integrate the graph structure into our feature encoder with multilevel sampling. This integration is beneficial because, on the one hand, the introduction of the graph structure enriches the local information. On the other hand, multilevel sampling captures high-level information due to the employment of a larger receptive field. Next, we define the graph structure as follows:
\begin{equation}
   \begin{aligned}
       E = \{(pi-pj||feature_i||feature_j)~|~\|p_i - p_j\|_2<r\}
   \end{aligned}
   \label{eq3}
\end{equation}
In Equation \ref{eq3}, $E$ denotes the edge set in the graph structure. We select the neighboring points around each point and use their distance and feature differences as the criteria for edge formation. $||$ is the contact operator. Following the methodology suggested by Kittenplon et al. \cite{Kittenplon_Eldar_Raviv_2020}, we encode edge features using three consecutive $setconv$ layers as our convolution mechanism and apply the furthest point sampling method for down-sampling.

In contrast to previous GNN methods, we introduce the concept of spatial memory in the feature extractor. The concept of memory is applied to the field of semantic segmentation \cite{tokmakov2017learning,nilsson2018semantic}, study shows that sequential input performs better than one-frame input, because sequential input allows the neural network to learn temporal information. The memory algorithm bases on the assumption of a certain level of temporal consistency from one frame to the next, previous work use GRU(Gate Recurrent Unit) to assist in semantic segmentation or other tasks by carrying prior information through the hidden layer.

Scene flow estimation carries certain temporal information, in this paper, the coarse flow is attached to point $P$ at time $t-1$, and the memory is retained to point $Q$ at time $t$ so as to learn the geometric information at different times.

\begin{equation}
    \begin{aligned}
        &p_{w,i} = p_{i}+f_{c,i} \\
        E' = \{(p_{w,i}-&qj||feature_i||feature_j)~|~\|p_{w,i} - q_j\|_2<r\}
        \label{eq4}
    \end{aligned}
\end{equation}

$p_{w,i}$ denotes the $i^{th}$ point in $P$ wrapped by coarse flow, $E'$ indicates the features of point $p_{w,i}$ with the MEMORY module, and we keep points to the next frame to learn the features of the surrounding points. As the previous operation, we also use $setconv$ layers to encode the features. $x$ and $x_{w,i}$ denotes the finished encoded feature.
\begin{equation}
    \begin{aligned}
        x_i = setconv(e_i),~e_i \in E \\
        x_{w,i} = setconv(e'_i),~e'_i \in E'
        \label{eq5}
    \end{aligned}
\end{equation}

\subsubsection{Correlation Matrix}
Neighbouring points share similar motion patterns, in the current studies, \cite{Kittenplon_Eldar_Raviv_2020,Wu_Wang_Li_Liu_Fuxin_2019} introduces smooth loss for penalising flow vectors that deviate too much from the surrounding points, and \cite{shi2022safit} also combines the knowledge of semantic segmentation with scene flow estimation model with the aim of segmenting objects in the scene that have the same motion model.

Flow consistency algorithms based on distance only can lead to prediction errors because the motion patterns of object boundary points are very different from those of surrounding points. So we model based on Euclidean distances and feature distances, the purpose of which is to reduce errors. Correlation Matrix is a module for learning point cloud similarity based on a feature encoder, where we combine the temporal and spatial features learned in the previous section to generate an $\hat{N} \times (N-N')$ matrix that measures the similarity between point levels.

\begin{equation}
    \begin{aligned}
    u_{i,n} &= (x^u_i||x^u_{w,i}) - (x^l_n||x^l_{w,i})\\
    g_{i,n} &= (p^u_i||p^u_{w,i}) - (p^l_n||p^l_{w,i})\\
    a_{i,n} &= MLP(u_{i,n})+MLP(g_{i,n}) 
    \end{aligned}
    \label{eq6}
\end{equation}
Where $a_{i,n}$ denotes the degree of similarity between the $i^{th}$ unlabelled point and the $n^{th}$ labelled point. Next, we assign each point $p_i$ a similarity vector, We map the similarity parameter to the interval $[0, 1]$ as a weight for flow up-sampling.
\begin{equation}
\begin{aligned}
   a_{i,t} &= softmax([a_{i,1}, a_{i,2}..., a_{i,N}])_n
    \label{eq7} 
\end{aligned}
\end{equation}

We update the scene flow vector with the correlation matrix that maps the ground truth labels to each point.
\begin{equation}
    f_i = \sum^{\hat{N}}_{n=1} a_{i,n}*\hat{f}_n
    \label{eq8}
\end{equation}
\subsection{Loss Function}
\label{sec:pagestyle}
Based on the pseudo label generation model in this paper, we propose a semi-supervised loss function (SSL PESUDO LABEL LOSS) to train the network based on the generated pseudo-label data. The loss function can be divided into the following 2 parts.

\subsubsection{Chamfer Loss}
chamfer loss measures the dissimilarity or discrepancy between two sets of points, often used for tasks like point cloud registration, shape matching, and generative modeling. Following \cite{Kittenplon_Eldar_Raviv_2020,Wu_Wang_Li_Liu_Fuxin_2019}, we calculate wrapped point by $\hat{P} = P+F$, and loss value. Equation shown in equation \ref{eq7}.
\begin{equation}
\begin{aligned}
    ChamferLoss(\hat{P}, Q) = \sum_{\hat{p} \in \hat{P}} \min_{q \in Q} \|\hat{p} - q\|^2 + \sum_{q \in Q} \min_{\hat{p} \in \hat{P}} \|q - \hat{p}\|^2&
    \label{eq9}
\end{aligned}
\end{equation}
\subsubsection{Weighted Smooth Loss}
Neighbouring point clouds share flow consistency, and the smooth loss is used to measure the degree of similarity with neighbouring points flow vectors. Smooth loss has been widely used in scene flow estimation \cite{Puy_Boulch_Marlet_2020,Wu_Wang_Li_Liu_Fuxin_2019}, however, assigning the same weight to all surrounding points may mentor the accumulation of streaming errors. 

We have implemented modifications to the smooth loss function. Unlike conventional methods that average the smoothness across all points, our modified approach preferentially assigns higher weights to points associated with ground truth labels. The objective of this adjustment is to more accurately aligneded the flow vectors of neighboring points with the ground truth labels, as elaborated in Equation \ref{eq8}. In this formulation, \( N \) and \( N' \) represent the sets of neighboring points, defined respectively as \( N_i = \{p_j \mid \|p^u_i - p_j\| < r\} \) and \( N'_i = \{p_j \mid \|p^l_i - p_j\| < r\} \). Here, \( r \) denotes the distance threshold. This nuanced modification is designed to enhance the precision of our model by ensuring a more focused and effective convergence of the flow vectors towards the ground truth.

In the loss function, we not only consider the flow consistency between points. We also compare the flow with the ground truth labels to prevent the error from widening further.

\begin{equation}
 \begin{aligned}
    WeightedSmoothLoss(P,F) = \beta_1\sum_{p^u_i\in P^u} \frac{1}{|N_i|} \sum_{p_j\in N_i} \|f_i-f_j\|_2 +\beta_2\sum_{p^l_i\in P^l} \frac{1}{|N'_i|} \sum_{p_j\in N'_i} \|f_i-f_j\|_2
    \label{eq10}
 \end{aligned}
\end{equation}

The overall loss function can be written as:

\begin{equation}
    \text{Loss}(P, F) = \alpha \cdot \text{ChamferLoss}(P, F) + \beta \cdot \text{WeightedSmoothLoss}(P, F)
    \label{eq11}
\end{equation}
\text{where} \(\alpha\) and \(\beta\) denote the weights of each loss function.

\begin{figure}[htbp]
	\centering
	\includegraphics[width=16.5cm]{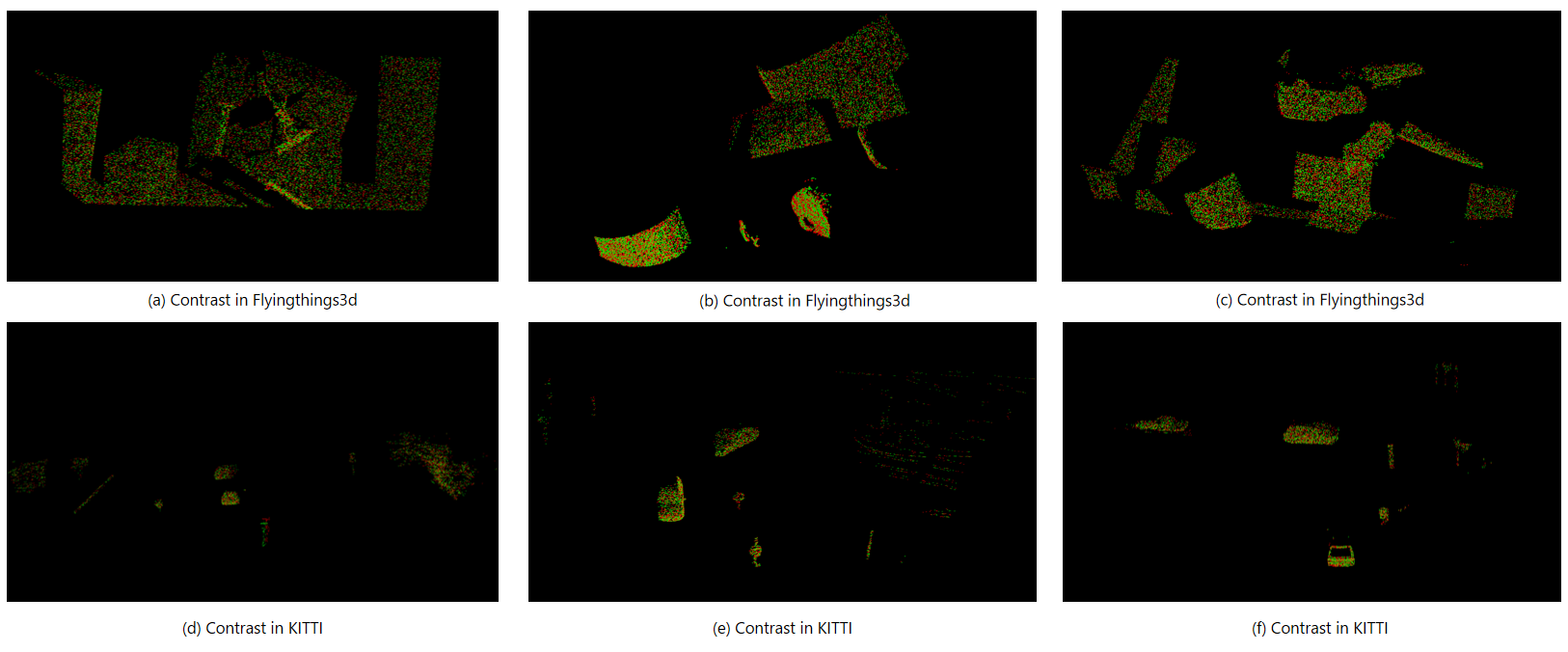}
	\caption{Results on FlyingThings3D (top) and KITTI (bottom). The red points represent the points of $Q$, while the green points denote the points wrapped through pseudo labels. The figure shows that our pseudo label generation model can approximate ground truth labels.}
    \label{fig4}
\end{figure}
\section{Experiment}
\label{sec:majhead}

In this section, we train and evaluate our SSLFlowNet on two datasets, FlyingThings3D \cite{mayer2016large} and KITTI Scene Flow \cite{Menze_Heipke_Geiger_2018}. Firstly, we compare with the current mainstream fully and semi-supervised models, as well as the pseudo-label generation model to illustrate the quality of our pseudo-label generation. Then, we combine the pseudo-label generation model with 3DFlownet \cite{wang2022matters} and take 1/8,1/16,1/32,1/64 in every batch for auxiliary training. Finally, we perform ablation experiments on different modules, and the experimental results prove that our proposed network structure can optimise the process of flow up-sampling. 

\subsection{Experimental Setting}
\textbf{Dataset} We conduct our experiments on FlyingThings3D and KITTI respectively. FlyingThings3D \cite{mayer2016large} is a synthetic dataset for optical flow, disparity and scene flow estimation. It consists of everyday objects flying along randomized 3D trajectories. It contains 19640 training examples and 3824 test examples. As most papers do, we only use one-quarter of the training data (4910 pairs). KITTI is a real-world scene flow dataset with 200 pairs for which 142 are used for testing without any fine-tuning. As most scene flow model do, we take the processing in HPLFLownet \cite{Gu_Wang_Wu_Lee_Wang_2019} to generate the non-occluded datasets. To be fair, we compare with other models on the same dataset as well as in training mode.

\textbf{Details} Our model is trained based on pytorch, using NVIDIA GeForce RTX 3090 as the hardware device. we train our model on synthetic FlyingThings3D training data and evaluate it on both FlyingThings3D test set and KITTI without finetune. Referring to most practices in the domain, we randomly take 8192 points per batch in training. In order to fit our semi-supervised model, we generate pseudo-labels by randomly sampling a certain percentage of these points as initial labels. We set the learning rate to 0.001 and the decay factor to 0.7, and decay in every 25 training epochs. The weights for chamfer loss and smooth loss are 0.75 and 0.25. We take Adam as the optimizer with default values for all parameters.

\textbf{Evaluation Metrics} We test our model with four evaluation metrics, including End Point Error (EPE), Accuracy Strict (AS), Accuracy Relax (AR), and Outliers (Out). We denote the estimated scene flow and ground truth scene flow as $F$ and $F_{gt}$, respectively. EPE(m): $\|F-F_{gt}\|_2$ averaged over all points. AS: the percentage of points whose EPE $<$0.05m or relative error$<$5\%. AR: the percentage of points whose EPE $<$0.1m or relative error $<$10\%. Out: the percentage of points whose EPE $>$ 0.3m or relative error $>$ 10\%.

\begin{figure}[htbp]
	\centering
	\includegraphics[width=16.5cm]{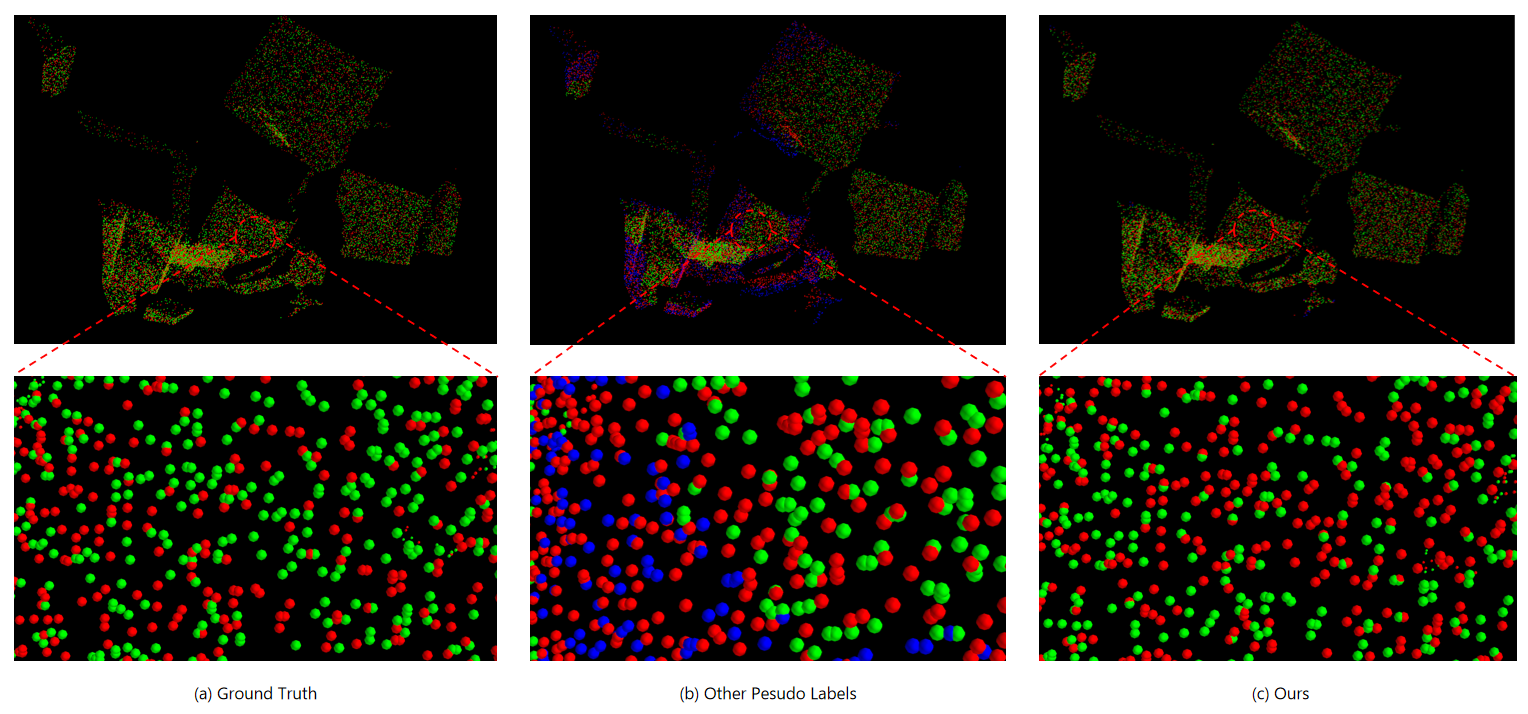}
	\caption{The results of our method compared with the ground truth label Fig. a, and the noisy label Fig. b are shown in abvoe. The red points shows the actual $Q$ , the green points shows the $P$ wrapped by generated labels, and the blue part shows the outliner points.}
    \label{fig5}
\end{figure}


\begin{figure}[h]
    \centering
    \begin{minipage}{0.45\textwidth}
        \centering
        \includegraphics[width=\linewidth]{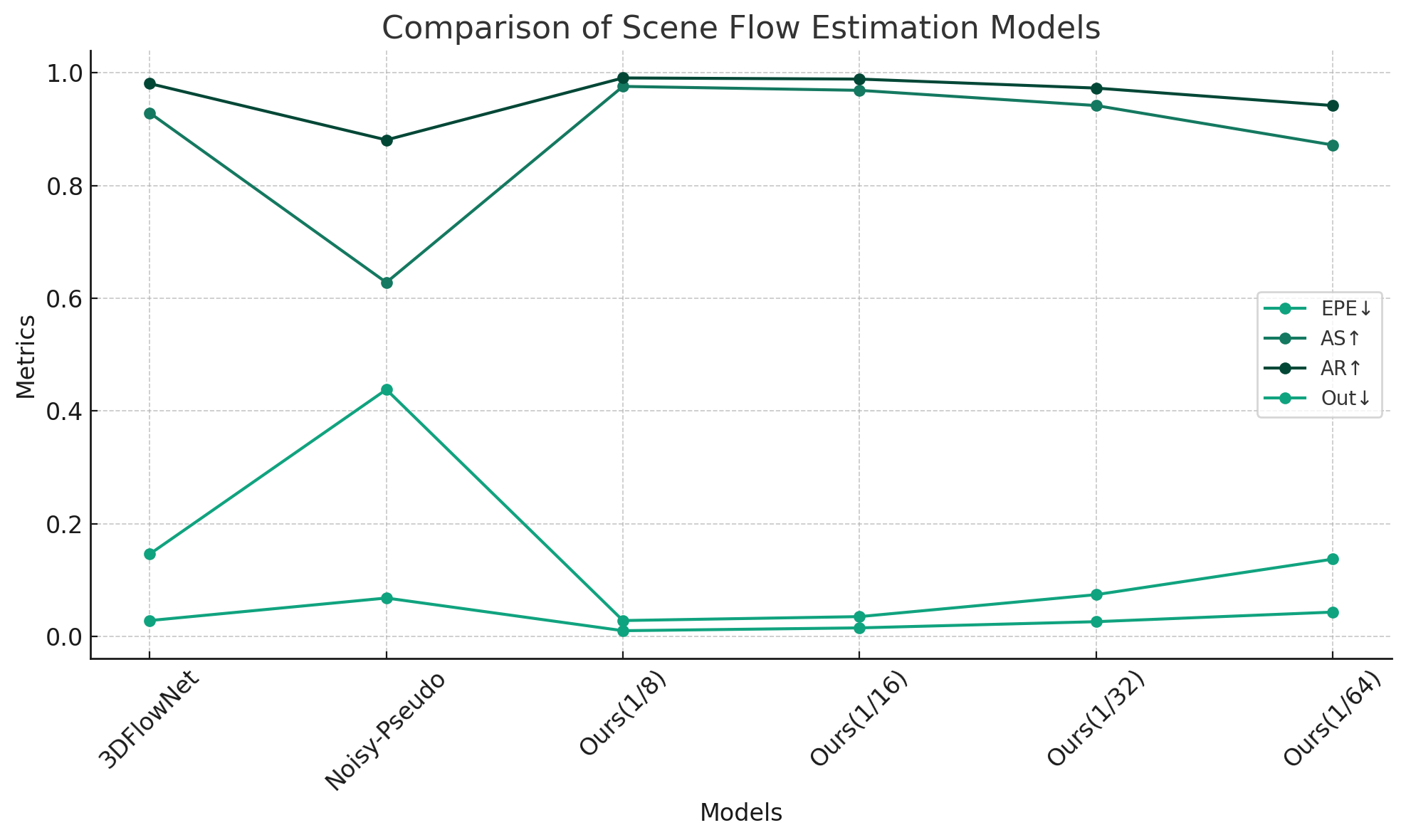}
        \caption{Comparative Analysis of Scene Flow Estimation Models with limited label generation in FlyingThings3D.}
        \label{fig:graphResult}
    \end{minipage}%
    \begin{minipage}{0.45\textwidth}

        This Figure. \ref{fig:graphResult} illustrates the comparative analysis of various scene flow estimation models. Each model is evaluated based on key metrics including End Point Error (EPE), Accuracy under threshold (AS), Accuracy Rate (AR), and the rate of Outliers (Out). The graph provides a visual representation of the performance of each model, highlighting the strengths and weaknesses in different aspects of scene flow estimation. We show our different ratios of the labeled data also makes the reasonable results. 
    \end{minipage}
\end{figure}

\begin{table}[!ht]
    \caption{Comparison of the generation quality of pseudo-labels with the current mainstream scenario flow prediction models.}
    \fontsize{7}{12}\selectfont
    \begin{minipage}[t]{0.48\linewidth}
    \begin{tabular}{l|l|l|l l l l}
    \hline
        Method & sup. & EPE↓ & AS↑ & AR↑ & Out↓  \\ \hline \hline
        \multicolumn{6}{c}{ \textbf{FlyingThings3D}} \\ \hline
        FlowNet3D & full & 0.114 & 0.413 & 0.77 & 0.602  \\ 
        PointPWC-Net & full & 0.059 & 0.738 & 0.928 & 0.342  \\  
        HLPFlownet & full & 0.08 & 0.614 & 0.856 & 0.429  \\ 
        3DFlowNet & full & 0.028 & 0.929 & 0.981 & 0.146  \\ \hline
        PointPWC-Net & self & 0.125 & 0.307 & 0.655 & 0.703  \\ 
        FlowStep3D & self & 0.085 & 0.536 & 0.826 & 0.420   \\ 
        RigidFlow & self & 0.069 & 0.596 & 0.871 & 0.464  \\ \hline
        Noisy-Pseudo & label & 0.068 & 0.628 & 0.881 & 0.438 \\
        \textbf{Ours(1/8)} & label & \textbf{0.010} & \textbf{0.976} & \textbf{0.991} & \textbf{0.028}\\
        \textbf{Ours(1/16)} & label & \textbf{0.015} & \textbf{0.969} & \textbf{0.989} & \textbf{0.035}\\
        \textbf{Ours(1/32)} & label & \textbf{0.026} & \textbf{0.942} & \textbf{0.973} & \textbf{0.074}\\ 
        \hline
        \end{tabular}
    \end{minipage}
    \begin{minipage}[t]{0.48\linewidth}
    \begin{tabular}{l|l|l|l l l l}
    \hline
        Method & sup. & EPE↓ & AS↑ & AR↑ & Out↓  \\ \hline \hline
        \multicolumn{6}{c}{ \textbf{KITTI}} \\ \hline
        FlowNet3D & full & 0.177& 0.374& 0.668& 0.527  \\ 
        PointPWC-Net & full & 0.069& 0.728& 0.888& 0.265  \\  
        HLPFlownet & full & 0.117 & 0.478 & 0.778& 0.410  \\ 
        3DFlowNet & full & 0.031 &0.905& 0.958 &0.161  \\ \hline
        PointPWC-Net & self & 0.069& 0.728& 0.888 &0.265  \\ 
        FlowStep3D & self & 0.055 &0.805& 0.925& 0.149   \\ 
        RigidFlow & self &0.062 &0.724 &0.892 &0.262  \\ \hline
        Noisy-Pseudo & label & 0.058 & 0.744 & 0.898 & 0.246 \\
        \textbf{Ours(1/8)} & label & \textbf{0.005} & \textbf{0.996} & \textbf{0.998} & \textbf{0.004}\\
        \textbf{Ours(1/16)} & label & \textbf{0.008} & \textbf{0.993} & \textbf{0.995} & \textbf{0.006}\\
        \textbf{Ours(1/32)} & label & \textbf{0.028} & \textbf{0.965} & \textbf{0.976} & \textbf{0.033}\\ 
        \hline
    \end{tabular}
    \end{minipage}
    \label{tb1}
\end{table}
\subsection{Evaluation of Pseudo Labels}
\label{ssec:subhead}
Our experiment results is shown in table \ref{tb1}. To reflect the generalisation ability of the model, we use random sampling to select 1/8,1/16,1/32,1/64 of the data. Subsequently, we adopt label propagation to spread the ground truth labels to the global. We adopt the same approach as \cite{iscen2019label} and use four metrics to measure the quality of our labels. To be fair, we exclude known labels when calculating EPE and accuracy. Our pseudo-label generation model was trained on Flyingthings for a total of 200 epochs, taking 10 hours. We tested the model on the KITTI dataset without fine-tuning.

\cite{li2021self} uses a consistent approach with \cite{Puy_Boulch_Marlet_2020} to generate labels, and \cite{iscen2019label} generate pseudo labels by optical flow learning. In contrast with these two, our label generation model is able to achieve higher accuracy. We are also able to generate high-quality pseudo-labels stably under different ground truth data volumes. Our results are shown in Figure \ref{fig4}.

\subsection{Results}
\label{sec:print}

We combine the generated pseudo-labels with 3DFlowNet \cite{Gu_Wang_Wu_Lee_Wang_2019} to show that training by pseudo-labelling is effective. We train on Flythings3D, test model accuracy on the Flythings3D test set and KITTI as shown in Table. \ref{tb2}, and compare it to the original benchmark model. The experimental results show that we sacrifice only a small portion of the accuracy, but greatly reduce the labelling cost.
\begin{table}[!ht]
    \caption{We use the generated pseudo-labels in conjunction with 3DFlowNet for training, and we use only 1/16 of the data for training.}
    \fontsize{6.5}{12}\selectfont
    \begin{minipage}[t]{0.5\linewidth}
    \begin{tabular}{l|l|l|l l l l}
    \hline
        Method & sup. & EPE↓ & AS↑ & AR↑ & Out↓  \\ \hline \hline
        \multicolumn{6}{c}{ \textbf{FlyingThings3D}} \\ \hline
        FlowNet3D & full & 0.114 & 0.413 & 0.77 & 0.602  \\ 
        PointPWC-Net & full & 0.059 & 0.738 & 0.928 & 0.342  \\  
        HLPFlownet & full & 0.08 & 0.614 & 0.856 & 0.429  \\ 
        \textbf{3DFlowNet} & full & \textbf{0.028} & \textbf{0.929} & \textbf{0.981} & \textbf{0.146}  \\ \hline
        PointPWC-Net & self & 0.125 & 0.307 & 0.655 & 0.703  \\ 
        FlowStep3D & self & 0.085 & 0.536 & 0.826 & 0.420   \\ 
        RigidFlow & self & 0.069 & 0.596 & 0.871 & 0.464  \\ \hline
        \textbf{Ours} & $1/8$ data & 0.045 & 0.842 & 0.951 & 0.195 \\
        \textbf{Ours} & $1/16$ data & 0.051 & 0.815 & 0.955 & 0.253 \\
        \textbf{Ours} & $1/32$ data & 0.059 & 0.747 & 0.892 & 0.318 \\
        \textbf{Ours} & $1/64$ data & 0.066 & 0.808 & 0.893 & 0.376 \\ \hline 
        \end{tabular}
        \end{minipage}
    \begin{minipage}[t]{0.5\linewidth}
    \begin{tabular}{l|l|l|l l l l}
    \hline
        Method & sup. & EPE↓ & AS↑ & AR↑ & Out↓  \\ \hline \hline
        \multicolumn{6}{c}{ \textbf{KITTI}} \\ \hline
        FlowNet3D & full & 0.177& 0.374& 0.668& 0.527  \\ 
        PointPWC-Net & full & 0.069& 0.728& 0.888& 0.265  \\  
        HLPFlownet & full & 0.117 & 0.478 & 0.778& 0.410  \\ 
        \textbf{3DFlowNet} & full & \textbf{0.031} &\textbf{0.905}& \textbf{0.958} &\textbf{0.161}  \\ \hline
        PointPWC-Net & self & 0.069& 0.728& 0.888 &0.265  \\ 
        FlowStep3D & self & 0.055 &0.805& 0.925& 0.149   \\ 
        RigidFlow & self &0.062 &0.724 &0.892 &0.262   \\ \hline
        \textbf{Ours} & $1/8$ data & 0.043 & 0.873 & 0.942 & 0.237 \\
        \textbf{Ours} & $1/16$ data & 0.052 & 0.810 & 0.949 & 0.306 \\
        \textbf{Ours} & $1/32$ data & 0.061 & 0.739 & 0.862 & 0.350 \\
        \textbf{Ours} & $1/64$ data & 0.075 & 0.683 & 0.822 & 0.415\\ \hline
    \end{tabular}
    \end{minipage}
    \label{tb2}
\end{table}

\subsection{Ablation Study}
We test the effectiveness of our individual modules on Flyingthings3D. We divide our network structure into several parts, up-sampling module based on the correlation matrix, spatial memory module, and smooth loss. The main idea of our experimental design is as follows. 

Firstly we take a simple up-sampling approach, i.e. we use the distance from the point to each labelled point as a weight, and perform a weighted average of the points as a first benchmark for our comparison without any refinement.

Next we explore the effects of spatial memory and improved smooth loss on our model. To prove the effectiveness of memory module, we remove the temporal features from \ref{eq5} and consider only the neighbourhood features of the current points. As seen from the results in the table, there is a dip in the experimental results, which is due to the fact that the feature changes due to the movement of the points were not taken into account when training the network.

For our weighted smooth loss experiments, to be fair, we take the averaged smooth loss used in \cite{Puy_Boulch_Marlet_2020} and \cite{Wu_Wang_Li_Liu_Fuxin_2019}. The result was shown in Table. \ref{tb3}.

\begin{table}[!ht]
    \centering
    \caption{We divided the network structure into 3 modules to test the effectiveness of our proposed methodology separately. It can be seen that the model performance decreases to some extent when different modules are deleted.}
    \fontsize{7}{12}\selectfont
    \begin{tabular}{c c c|l l l l}
    \hline
        Correlation Matrix & Spatial Memory & Weighted Smooth Loss & EPE↓ & AS↑ & AR↑ & Out↓  \\ \hline \hline
         &  &  & 0.297 & 0.142 & 0.323 & 0.914  \\ 
        \ding{51} &  &  & 0.062 & 0.738 & 0.874 & 0.223  \\  
        \ding{51} & \ding{51} &  & 0.056 & 0.815 & 0.917 & 0.193  \\ 
        \ding{51} &  & \ding{51} & 0.049 & 0.850 & 0.933 & 0.162   \\
        \ding{51} & \ding{51} & \ding{51} & 0.043 & 0.872 & 0.942 & 0.137 \\ \hline
        \end{tabular}
    \label{tb3}
\end{table}

\section{Discussion}
In this section, we discuss in detail the problems associated with point cloud scene flow estimation based on pseudo-label training. This includes the limitations of the current algorithms. We will also further propose problems that still need to be solved for future research.


\begin{figure}[htbp]
	\centering
	\includegraphics[width=16.5cm]{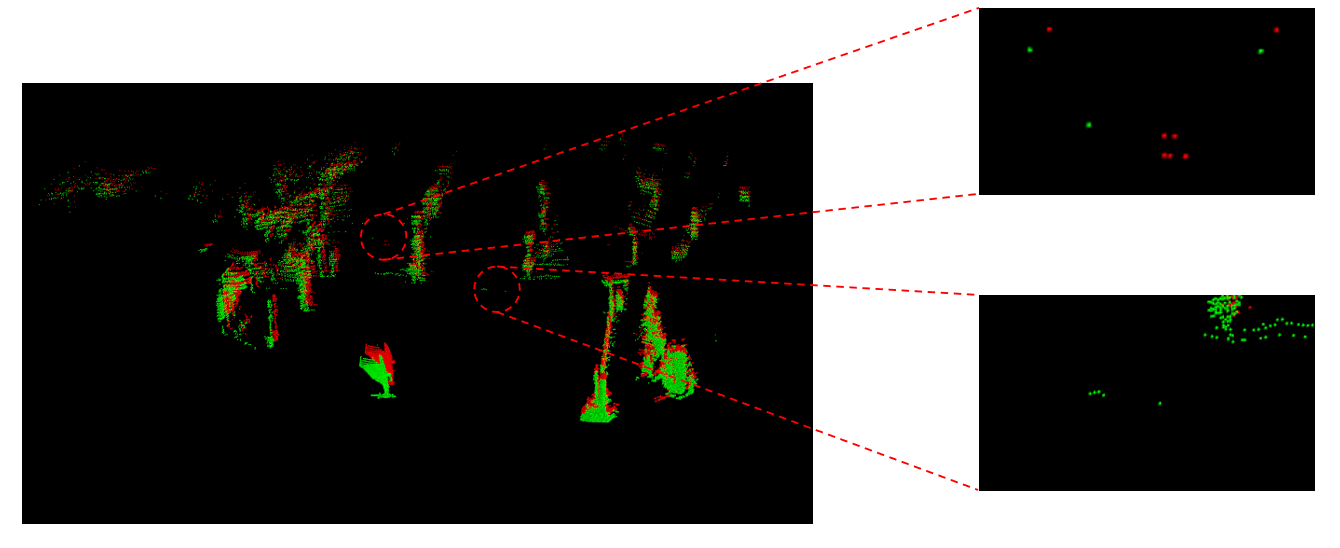}
	\caption{The noise points in the KITTI dataset are shown in the figure, with the red points indicating the $t-1$ moment in the scene and the green points indicating the $t$ moment in the scene. The zoom area are noise points, and the presence of noise points will have a certain impact on the accuracy of the scene flow.}
    \label{fig6}
\end{figure}

 \textbf{Noise in Point Clouds:} Point clouds inherently possess noise or outliers, often attributable to sensor inaccuracies or environmental factors. The noise, an inevitable byproduct of the scanning and reconstruction processes, poses significant challenges in feature extraction and can mislead the search for corresponding points. While several approaches \cite{rakotosaona2020pointcleannet, luo2020differentiable, luo2021score} have explored denoising via neural networks, effectively mitigating the errors induced by noise points within the context of scene flow estimation remains an open problem.

Predominantly, scene flow models are trained using the simulated dataset FlyingThings3D. However, when these models are transitioned to real-world scenarios such as those represented by KITTI, they exhibit varying degrees of performance degradation. Li et al. \cite{li2022learning} proposed a noisy-label-aware training scheme that leverages the geometric relations of points, yet this approach continues to grapple with limitations in accuracy.

\textbf{Occlusions and Scene Flow:} Occlusions, resulting from the time interval between two frames, can lead to the disappearance of objects and significantly impact flow estimation accuracy. Recent research efforts \cite{saxena2019pwoc, ouyang2021occlusion, zanfir2015large} have focused on developing methodologies to address occlusions. Notably, Ouyang et al. \cite{ouyang2021occlusion} introduced a novel framework to assess occlusion in 3D scene flow estimation on point clouds. However, the accuracy of unsupervised and semi-supervised models in predicting occlusions remains suboptimal.

Current models frequently employ the data processing strategy outlined in HPLFlowNet \cite{Gu_Wang_Wu_Lee_Wang_2019}, which simplifies the flow estimation task by eliminating occluded points. This, however, can adversely affect model generalization. Thus, the development of effective pseudo-labels for training models, with the aim of enhancing generalization, presents a promising avenue for future research.

\section{Conclusion}
\label{sec:page}

In this paper, we introduce a pseudo-label generation model predicated on ground truth labels. This model is grounded in the time-space feature analysis and the assumption of motion consistency within point clouds. To upscale the labels, we employ both chamfer loss and an enhanced smoothing loss technique. Empirical evidence from our experiments demonstrates that the pseudo-labels generated by our model exhibit state-of-the-art performance and are amenable to integration with various supervised methodologies. Significantly, our approach reduces the reliance on manual labeling and facilitates the training of high-caliber models even when limited data labels are available.

\section{Acknowledge}
The paper is supported by the Natural Science Foundation of China (No. 62072388), Collaborative Project fund of Fuzhou-Xiamen-Quanzhou Innovation Zone(No.3502ZCQXT202001),the industry guidance project foundation of science technology bureau of Fujian province in 2020(No.2020H0047), , and Fujian Sunshine Charity Foundation.
\bibliography{mybibfile}

\begin{thebibliography}{35}
\expandafter\ifx\csname natexlab\endcsname\relax\def\natexlab#1{#1}\fi
\providecommand{\url}[1]{\texttt{#1}}
\providecommand{\href}[2]{#2}
\providecommand{\path}[1]{#1}
\providecommand{\DOIprefix}{doi:}
\providecommand{\ArXivprefix}{arXiv:}
\providecommand{\URLprefix}{URL: }
\providecommand{\Pubmedprefix}{pmid:}
\providecommand{\doi}[1]{\href{http://dx.doi.org/#1}{\path{#1}}}
\providecommand{\Pubmed}[1]{\href{pmid:#1}{\path{#1}}}
\providecommand{\bibinfo}[2]{#2}
\ifx\xfnm\relax \def\xfnm[#1]{\unskip,\space#1}\fi
\bibitem[{Dong et~al.(2022)Dong, Zhang, Li, Sun and Xiong}]{dong2022exploiting}
\bibinfo{author}{Dong, G.}, \bibinfo{author}{Zhang, Y.}, \bibinfo{author}{Li, H.}, \bibinfo{author}{Sun, X.}, \bibinfo{author}{Xiong, Z.}, \bibinfo{year}{2022}.
\newblock \bibinfo{title}{Exploiting rigidity constraints for lidar scene flow estimation}, in: \bibinfo{booktitle}{Proceedings of the IEEE/CVF Conference on Computer Vision and Pattern Recognition}, pp. \bibinfo{pages}{12776--12785}.
\bibitem[{Gu et~al.(2019)Gu, Wang, Wu, Lee and Wang}]{Gu_Wang_Wu_Lee_Wang_2019}
\bibinfo{author}{Gu, X.}, \bibinfo{author}{Wang, Y.}, \bibinfo{author}{Wu, C.}, \bibinfo{author}{Lee, Y.}, \bibinfo{author}{Wang, P.}, \bibinfo{year}{2019}.
\newblock \bibinfo{title}{Hplflownet: Hierarchical permutohedral lattice flownet for scene flow estimation on large-scale point clouds}.
\newblock \bibinfo{journal}{Cornell University - arXiv,Cornell University - arXiv} .
\bibitem[{Huguet and Devernay(2007)}]{huguet2007variational}
\bibinfo{author}{Huguet, F.}, \bibinfo{author}{Devernay, F.}, \bibinfo{year}{2007}.
\newblock \bibinfo{title}{A variational method for scene flow estimation from stereo sequences}, in: \bibinfo{booktitle}{2007 IEEE 11th International Conference on Computer Vision}, \bibinfo{organization}{IEEE}. pp. \bibinfo{pages}{1--7}.
\bibitem[{Iscen et~al.(2019)Iscen, Tolias, Avrithis and Chum}]{iscen2019label}
\bibinfo{author}{Iscen, A.}, \bibinfo{author}{Tolias, G.}, \bibinfo{author}{Avrithis, Y.}, \bibinfo{author}{Chum, O.}, \bibinfo{year}{2019}.
\newblock \bibinfo{title}{Label propagation for deep semi-supervised learning}, in: \bibinfo{booktitle}{Proceedings of the IEEE/CVF conference on computer vision and pattern recognition}, pp. \bibinfo{pages}{5070--5079}.
\bibitem[{Kittenplon et~al.(2020)Kittenplon, Eldar and Raviv}]{Kittenplon_Eldar_Raviv_2020}
\bibinfo{author}{Kittenplon, Y.}, \bibinfo{author}{Eldar, Y.}, \bibinfo{author}{Raviv, D.}, \bibinfo{year}{2020}.
\newblock \bibinfo{title}{Flowstep3d: Model unrolling for self-supervised scene flow estimation}.
\newblock \bibinfo{journal}{Cornell University - arXiv,Cornell University - arXiv} .
\bibitem[{Lee et~al.(2013)}]{lee2013pseudo}
\bibinfo{author}{Lee, D.H.}, et~al., \bibinfo{year}{2013}.
\newblock \bibinfo{title}{Pseudo-label: The simple and efficient semi-supervised learning method for deep neural networks}, in: \bibinfo{booktitle}{Workshop on challenges in representation learning, ICML}, \bibinfo{organization}{Atlanta}. p. \bibinfo{pages}{896}.
\bibitem[{Li et~al.(2022a)Li, Zheng, Li and Ghanem}]{li2022learning}
\bibinfo{author}{Li, B.}, \bibinfo{author}{Zheng, C.}, \bibinfo{author}{Li, G.}, \bibinfo{author}{Ghanem, B.}, \bibinfo{year}{2022}a.
\newblock \bibinfo{title}{Learning scene flow in 3d point clouds with noisy pseudo labels}.
\newblock \bibinfo{journal}{arXiv preprint arXiv:2203.12655} .
\bibitem[{Li et~al.(2023)Li, Casas and Urtasun}]{li2023memoryseg}
\bibinfo{author}{Li, E.}, \bibinfo{author}{Casas, S.}, \bibinfo{author}{Urtasun, R.}, \bibinfo{year}{2023}.
\newblock \bibinfo{title}{Memoryseg: Online lidar semantic segmentation with a latent memory}, in: \bibinfo{booktitle}{Proceedings of the IEEE/CVF International Conference on Computer Vision}, pp. \bibinfo{pages}{745--754}.
\bibitem[{Li(2023)}]{li2023edge}
\bibinfo{author}{Li, L.}, \bibinfo{year}{2023}.
\newblock \bibinfo{title}{Edge aware learning for 3d point cloud}.
\newblock \bibinfo{journal}{arXiv preprint arXiv:2309.13472} .
\bibitem[{Li et~al.(2021)Li, Lin and Xie}]{li2021self}
\bibinfo{author}{Li, R.}, \bibinfo{author}{Lin, G.}, \bibinfo{author}{Xie, L.}, \bibinfo{year}{2021}.
\newblock \bibinfo{title}{Self-point-flow: Self-supervised scene flow estimation from point clouds with optimal transport and random walk}, in: \bibinfo{booktitle}{Proceedings of the IEEE/CVF conference on computer vision and pattern recognition}, pp. \bibinfo{pages}{15577--15586}.
\bibitem[{Li et~al.(2022b)Li, Zhang, Lin, Wang and Shen}]{li2022rigidflow}
\bibinfo{author}{Li, R.}, \bibinfo{author}{Zhang, C.}, \bibinfo{author}{Lin, G.}, \bibinfo{author}{Wang, Z.}, \bibinfo{author}{Shen, C.}, \bibinfo{year}{2022}b.
\newblock \bibinfo{title}{Rigidflow: Self-supervised scene flow learning on point clouds by local rigidity prior}, in: \bibinfo{booktitle}{Proceedings of the IEEE/CVF Conference on Computer Vision and Pattern Recognition}, pp. \bibinfo{pages}{16959--16968}.
\bibitem[{Liu et~al.(2019)Liu, Qi and Guibas}]{liu2019flownet3d}
\bibinfo{author}{Liu, X.}, \bibinfo{author}{Qi, C.R.}, \bibinfo{author}{Guibas, L.J.}, \bibinfo{year}{2019}.
\newblock \bibinfo{title}{Flownet3d: Learning scene flow in 3d point clouds}, in: \bibinfo{booktitle}{Proceedings of the IEEE/CVF conference on computer vision and pattern recognition}, pp. \bibinfo{pages}{529--537}.
\bibitem[{Luo and Hu(2020)}]{luo2020differentiable}
\bibinfo{author}{Luo, S.}, \bibinfo{author}{Hu, W.}, \bibinfo{year}{2020}.
\newblock \bibinfo{title}{Differentiable manifold reconstruction for point cloud denoising}, in: \bibinfo{booktitle}{Proceedings of the 28th ACM international conference on multimedia}, pp. \bibinfo{pages}{1330--1338}.
\bibitem[{Luo and Hu(2021)}]{luo2021score}
\bibinfo{author}{Luo, S.}, \bibinfo{author}{Hu, W.}, \bibinfo{year}{2021}.
\newblock \bibinfo{title}{Score-based point cloud denoising (learning implicit gradient fields for point cloud denoising)}.
\newblock \bibinfo{journal}{arXiv e-prints} , \bibinfo{pages}{arXiv--2107}.
\bibitem[{Mayer et~al.(2016)Mayer, Ilg, Hausser, Fischer, Cremers, Dosovitskiy and Brox}]{mayer2016large}
\bibinfo{author}{Mayer, N.}, \bibinfo{author}{Ilg, E.}, \bibinfo{author}{Hausser, P.}, \bibinfo{author}{Fischer, P.}, \bibinfo{author}{Cremers, D.}, \bibinfo{author}{Dosovitskiy, A.}, \bibinfo{author}{Brox, T.}, \bibinfo{year}{2016}.
\newblock \bibinfo{title}{A large dataset to train convolutional networks for disparity, optical flow, and scene flow estimation}, in: \bibinfo{booktitle}{Proceedings of the IEEE conference on computer vision and pattern recognition}, pp. \bibinfo{pages}{4040--4048}.
\bibitem[{Menze and Geiger(2015)}]{menze2015object}
\bibinfo{author}{Menze, M.}, \bibinfo{author}{Geiger, A.}, \bibinfo{year}{2015}.
\newblock \bibinfo{title}{Object scene flow for autonomous vehicles}, in: \bibinfo{booktitle}{Proceedings of the IEEE conference on computer vision and pattern recognition}, pp. \bibinfo{pages}{3061--3070}.
\bibitem[{Menze et~al.(2018)Menze, Heipke and Geiger}]{Menze_Heipke_Geiger_2018}
\bibinfo{author}{Menze, M.}, \bibinfo{author}{Heipke, C.}, \bibinfo{author}{Geiger, A.}, \bibinfo{year}{2018}.
\newblock \bibinfo{title}{Object scene flow}.
\newblock \bibinfo{journal}{ISPRS Journal of Photogrammetry and Remote Sensing} , \bibinfo{pages}{60–76}.
\bibitem[{Mittal et~al.(2019)Mittal, Okorn and Held}]{Mittal_Okorn_Held_2019}
\bibinfo{author}{Mittal, H.}, \bibinfo{author}{Okorn, B.}, \bibinfo{author}{Held, D.}, \bibinfo{year}{2019}.
\newblock \bibinfo{title}{Just go with the flow: Self-supervised scene flow estimation}.
\newblock \bibinfo{journal}{arXiv: Computer Vision and Pattern Recognition,arXiv: Computer Vision and Pattern Recognition} .
\bibitem[{Nilsson and Sminchisescu(2018)}]{nilsson2018semantic}
\bibinfo{author}{Nilsson, D.}, \bibinfo{author}{Sminchisescu, C.}, \bibinfo{year}{2018}.
\newblock \bibinfo{title}{Semantic video segmentation by gated recurrent flow propagation}, in: \bibinfo{booktitle}{Proceedings of the IEEE conference on computer vision and pattern recognition}, pp. \bibinfo{pages}{6819--6828}.
\bibitem[{Ouyang and Raviv(2021)}]{ouyang2021occlusion}
\bibinfo{author}{Ouyang, B.}, \bibinfo{author}{Raviv, D.}, \bibinfo{year}{2021}.
\newblock \bibinfo{title}{Occlusion guided self-supervised scene flow estimation on 3d point clouds}, in: \bibinfo{booktitle}{2021 International Conference on 3D Vision (3DV)}, \bibinfo{organization}{IEEE}. pp. \bibinfo{pages}{782--791}.
\bibitem[{Paul et~al.(2021)Paul, Danelljan, Van~Gool and Timofte}]{paul2021local}
\bibinfo{author}{Paul, M.}, \bibinfo{author}{Danelljan, M.}, \bibinfo{author}{Van~Gool, L.}, \bibinfo{author}{Timofte, R.}, \bibinfo{year}{2021}.
\newblock \bibinfo{title}{Local memory attention for fast video semantic segmentation}, in: \bibinfo{booktitle}{2021 IEEE/RSJ International Conference on Intelligent Robots and Systems (IROS)}, \bibinfo{organization}{IEEE}. pp. \bibinfo{pages}{1102--1109}.
\bibitem[{Puy et~al.(2020)Puy, Boulch and Marlet}]{Puy_Boulch_Marlet_2020}
\bibinfo{author}{Puy, G.}, \bibinfo{author}{Boulch, A.}, \bibinfo{author}{Marlet, R.}, \bibinfo{year}{2020}.
\newblock \bibinfo{title}{Flot: Scene flow on point clouds guided by optimal transport}.
\newblock \bibinfo{journal}{Cornell University - arXiv,Cornell University - arXiv} .
\bibitem[{Qi et~al.(2017a)Qi, Su, Mo and Guibas}]{qi2017pointnet}
\bibinfo{author}{Qi, C.R.}, \bibinfo{author}{Su, H.}, \bibinfo{author}{Mo, K.}, \bibinfo{author}{Guibas, L.J.}, \bibinfo{year}{2017}a.
\newblock \bibinfo{title}{Pointnet: Deep learning on point sets for 3d classification and segmentation}, in: \bibinfo{booktitle}{Proceedings of the IEEE conference on computer vision and pattern recognition}, pp. \bibinfo{pages}{652--660}.
\bibitem[{Qi et~al.(2017b)Qi, Yi, Su and Guibas}]{qi2017pointnet++}
\bibinfo{author}{Qi, C.R.}, \bibinfo{author}{Yi, L.}, \bibinfo{author}{Su, H.}, \bibinfo{author}{Guibas, L.J.}, \bibinfo{year}{2017}b.
\newblock \bibinfo{title}{Pointnet++: Deep hierarchical feature learning on point sets in a metric space}.
\newblock \bibinfo{journal}{Advances in neural information processing systems} \bibinfo{volume}{30}.
\bibitem[{Rakotosaona et~al.(2020)Rakotosaona, La~Barbera, Guerrero, Mitra and Ovsjanikov}]{rakotosaona2020pointcleannet}
\bibinfo{author}{Rakotosaona, M.J.}, \bibinfo{author}{La~Barbera, V.}, \bibinfo{author}{Guerrero, P.}, \bibinfo{author}{Mitra, N.J.}, \bibinfo{author}{Ovsjanikov, M.}, \bibinfo{year}{2020}.
\newblock \bibinfo{title}{Pointcleannet: Learning to denoise and remove outliers from dense point clouds}, in: \bibinfo{booktitle}{Computer graphics forum}, \bibinfo{organization}{Wiley Online Library}. pp. \bibinfo{pages}{185--203}.
\bibitem[{Saxena et~al.(2019)Saxena, Schuster, Wasenmuller and Stricker}]{saxena2019pwoc}
\bibinfo{author}{Saxena, R.}, \bibinfo{author}{Schuster, R.}, \bibinfo{author}{Wasenmuller, O.}, \bibinfo{author}{Stricker, D.}, \bibinfo{year}{2019}.
\newblock \bibinfo{title}{Pwoc-3d: Deep occlusion-aware end-to-end scene flow estimation}, in: \bibinfo{booktitle}{2019 IEEE Intelligent Vehicles Symposium (IV)}, \bibinfo{organization}{IEEE}. pp. \bibinfo{pages}{324--331}.
\bibitem[{Shen et~al.(2023)Shen, Hui, Xie and Yang}]{shen2023self}
\bibinfo{author}{Shen, Y.}, \bibinfo{author}{Hui, L.}, \bibinfo{author}{Xie, J.}, \bibinfo{author}{Yang, J.}, \bibinfo{year}{2023}.
\newblock \bibinfo{title}{Self-supervised 3d scene flow estimation guided by superpoints}, in: \bibinfo{booktitle}{Proceedings of the IEEE/CVF Conference on Computer Vision and Pattern Recognition}, pp. \bibinfo{pages}{5271--5280}.
\bibitem[{Shi and Rajkumar(2020)}]{shi2020point}
\bibinfo{author}{Shi, W.}, \bibinfo{author}{Rajkumar, R.}, \bibinfo{year}{2020}.
\newblock \bibinfo{title}{Point-gnn: Graph neural network for 3d object detection in a point cloud}, in: \bibinfo{booktitle}{Proceedings of the IEEE/CVF conference on computer vision and pattern recognition}, pp. \bibinfo{pages}{1711--1719}.
\bibitem[{Shi and Ma(2022)}]{shi2022safit}
\bibinfo{author}{Shi, Y.}, \bibinfo{author}{Ma, K.}, \bibinfo{year}{2022}.
\newblock \bibinfo{title}{Safit: Segmentation-aware scene flow with improved transformer}, in: \bibinfo{booktitle}{2022 International Conference on Robotics and Automation (ICRA)}, \bibinfo{organization}{IEEE}. pp. \bibinfo{pages}{10648--10655}.
\bibitem[{Tokmakov et~al.(2017)Tokmakov, Alahari and Schmid}]{tokmakov2017learning}
\bibinfo{author}{Tokmakov, P.}, \bibinfo{author}{Alahari, K.}, \bibinfo{author}{Schmid, C.}, \bibinfo{year}{2017}.
\newblock \bibinfo{title}{Learning video object segmentation with visual memory}, in: \bibinfo{booktitle}{Proceedings of the IEEE International Conference on Computer Vision}, pp. \bibinfo{pages}{4481--4490}.
\bibitem[{Vedula et~al.(1999)Vedula, Baker, Rander, Collins and Kanade}]{vedula1999three}
\bibinfo{author}{Vedula, S.}, \bibinfo{author}{Baker, S.}, \bibinfo{author}{Rander, P.}, \bibinfo{author}{Collins, R.}, \bibinfo{author}{Kanade, T.}, \bibinfo{year}{1999}.
\newblock \bibinfo{title}{Three-dimensional scene flow}, in: \bibinfo{booktitle}{Proceedings of the Seventh IEEE International Conference on Computer Vision}, \bibinfo{organization}{IEEE}. pp. \bibinfo{pages}{722--729}.
\bibitem[{Vogel et~al.(2011)Vogel, Schindler and Roth}]{vogel20113d}
\bibinfo{author}{Vogel, C.}, \bibinfo{author}{Schindler, K.}, \bibinfo{author}{Roth, S.}, \bibinfo{year}{2011}.
\newblock \bibinfo{title}{3d scene flow estimation with a rigid motion prior}, in: \bibinfo{booktitle}{2011 International Conference on Computer Vision}, \bibinfo{organization}{IEEE}. pp. \bibinfo{pages}{1291--1298}.
\bibitem[{Wang et~al.(2022)Wang, Hu, Liu, Zhou, Tomizuka, Zhan and Wang}]{wang2022matters}
\bibinfo{author}{Wang, G.}, \bibinfo{author}{Hu, Y.}, \bibinfo{author}{Liu, Z.}, \bibinfo{author}{Zhou, Y.}, \bibinfo{author}{Tomizuka, M.}, \bibinfo{author}{Zhan, W.}, \bibinfo{author}{Wang, H.}, \bibinfo{year}{2022}.
\newblock \bibinfo{title}{What matters for 3d scene flow network}, in: \bibinfo{booktitle}{European Conference on Computer Vision}, \bibinfo{organization}{Springer}. pp. \bibinfo{pages}{38--55}.
\bibitem[{Wu et~al.(2019)Wu, Wang, Li, Liu and Fuxin}]{Wu_Wang_Li_Liu_Fuxin_2019}
\bibinfo{author}{Wu, W.}, \bibinfo{author}{Wang, Z.}, \bibinfo{author}{Li, Z.}, \bibinfo{author}{Liu, W.}, \bibinfo{author}{Fuxin, L.}, \bibinfo{year}{2019}.
\newblock \bibinfo{title}{Pointpwc-net: A coarse-to-fine network for supervised and self-supervised scene flow estimation on 3d point clouds}.
\newblock \bibinfo{journal}{Cornell University - arXiv,Cornell University - arXiv} .
\bibitem[{Zanfir and Sminchisescu(2015)}]{zanfir2015large}
\bibinfo{author}{Zanfir, A.}, \bibinfo{author}{Sminchisescu, C.}, \bibinfo{year}{2015}.
\newblock \bibinfo{title}{Large displacement 3d scene flow with occlusion reasoning}, in: \bibinfo{booktitle}{Proceedings of the IEEE International Conference on Computer Vision}, pp. \bibinfo{pages}{4417--4425}.

\end{thebibliography}

\end{document}